%% file: OymakSPL.tex
\documentclass[journal]{IEEEtran}

\ifCLASSINFOpdf
\else
   \usepackage[dvips]{graphicx}
\fi
\usepackage{url}

\usepackage{graphicx}
\input{notation}

\begin{document}

\title{\red{Provable} Super-Convergence with \red{a Large\\Cyclical} Learning Rate}

\author{Samet Oymak \IEEEmembership{Member, IEEE}
\thanks{Submitted on April 6, 2021. This work was supported in part by the NSF under CNS grant 1932254 and the CAREER award 2046816. }}

\markboth{Journal of \LaTeX\ Class Files, Vol. 14, No. 8, August 2015}
{Shell \MakeLowercase{\textit{et al.}}: Bare Demo of IEEEtran.cls for IEEE Journals}
\maketitle

\begin{abstract}
Conventional wisdom dictates that learning rate should be in the stable regime so that gradient-based algorithms don't blow up. This letter introduces a simple scenario where an \red{unstably large} learning rate scheme leads to a super fast convergence, with the convergence rate depending only logarithmically on the condition number of the problem. Our scheme uses a Cyclical Learning Rate where we periodically take one large unstable step and several small stable steps to compensate for the instability. These findings also help explain the empirical observations of [Smith and Topin, 2019] where they \red{show that} CLR with a large maximum learning rate \red{can dramatically accelerate learning and lead to so-called} ``super-convergence''. We prove that our scheme excels in the problems where Hessian exhibits a bimodal spectrum and the eigenvalues can be grouped into two clusters (small and large). The \red{unstably large} step is the key to enabling fast convergence over the small eigen-spectrum.

\end{abstract}

\begin{IEEEkeywords}
Convergence of numerical methods, Iterative algorithms, Gradient methods
\end{IEEEkeywords}

\IEEEpeerreviewmaketitle

\section{Introduction}

Consider a least-squares problem with design matrix $\X\in\R^{n\times p}$ and labels $\y\in\R^{n}$. We wish to solve for
\[
\bts=\arg\min_{\bt\in\R^p}\frac{1}{2}\tn{\y-\X\bt}^2.
\] 
If we use a gradient-based algorithm the rate of convergence obviously depends on the condition number $\kappa$ of $\X$. Here $\kappa=L/\mu$ where the smoothness $L$ and strong convexity $\mu$ of the problem is given by the maximum and minimum eigenvalues of the Hessian matrix $\X^\top\X$ as follows
\[
L=\|\X^\top\X\|,\quad\mu=\smn{\X^\top\X}.
\]
Here, $\smn{},\|\cdot\|$ denote the smallest/largest singular value of a matrix respectively. \red{Standard} gradient descent (GD) requires $\kappa\log(\eps^{-1})$ iterations to achieve $\eps$-accuracy. Nesterov's acceleration can improve this to $\sqrt{\kappa}\log(\eps^{-1})$. In general, consider the iterations
\[
\bt_{t+1}=\bt_t-\eta_t\X^\top(\y-\X\bt_t).
\]
Here, the contraction matrix $\Cb_t=\Iden-\eta_t\X^T\X$ governs the rate of convergence. Over the $i$th eigen-direction of $\X^\top\X$ with eigenvalue $\la_i$, the convergence/contraction rate is given by $1-\eta_t\la_i$. Setting a fixed stable learning rate of $\eta_t=1/L$, the issue is that gradient descent optimizes small eigen-directions much slower than the large eigen-directions. Faster learning over small eigen-directions can be facilitated by a large learning rate so that $1-\eta_t\la_i$ is closer to $0$ even for small $\la_i$'s. However this would lead to instability i.e.~$\|\Cb_t\|>1$.

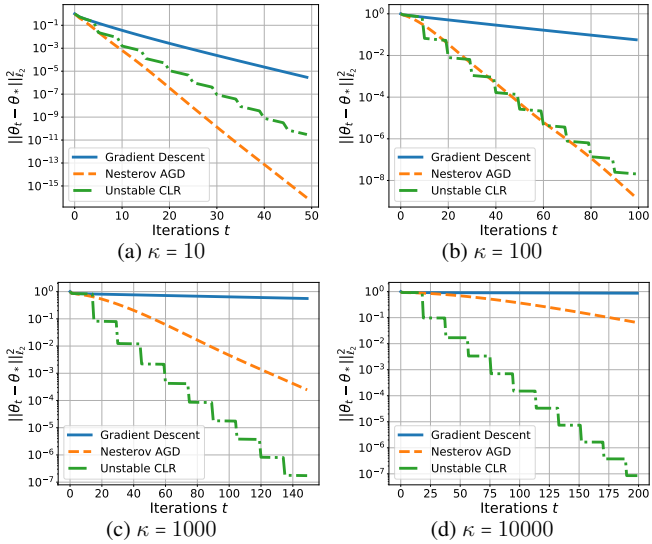
\begin{figure}
\begin{tikzpicture}
\node at (0,0){\includegraphics[width=0.24\textwidth]{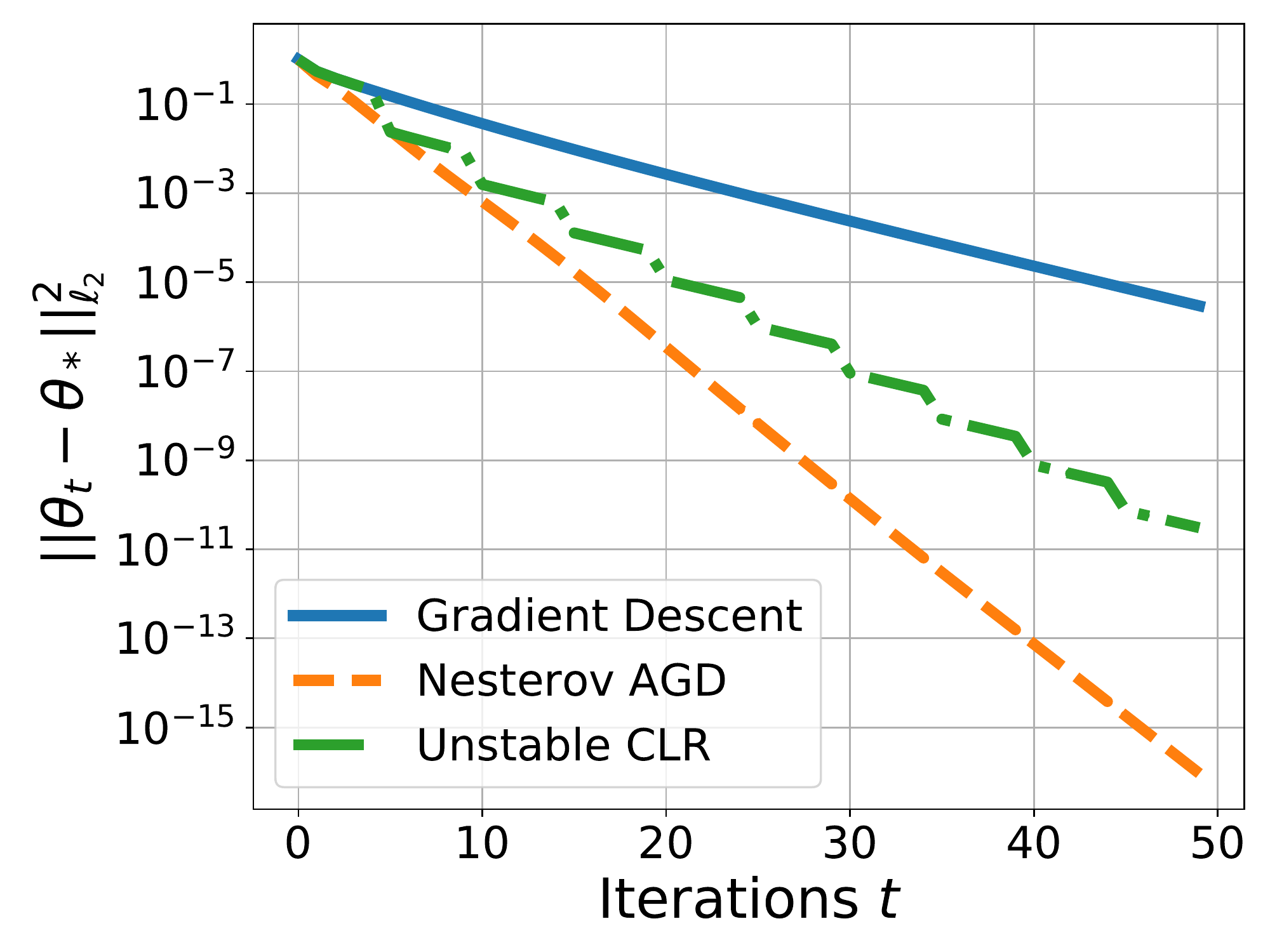}};
\node at (0,-1.75) [scale=0.6] {\Large{(a) $\kappa=10$}};
\end{tikzpicture}~\hspace{-10pt}
\begin{tikzpicture}
\node at (0,0){\includegraphics[width=0.24\textwidth]{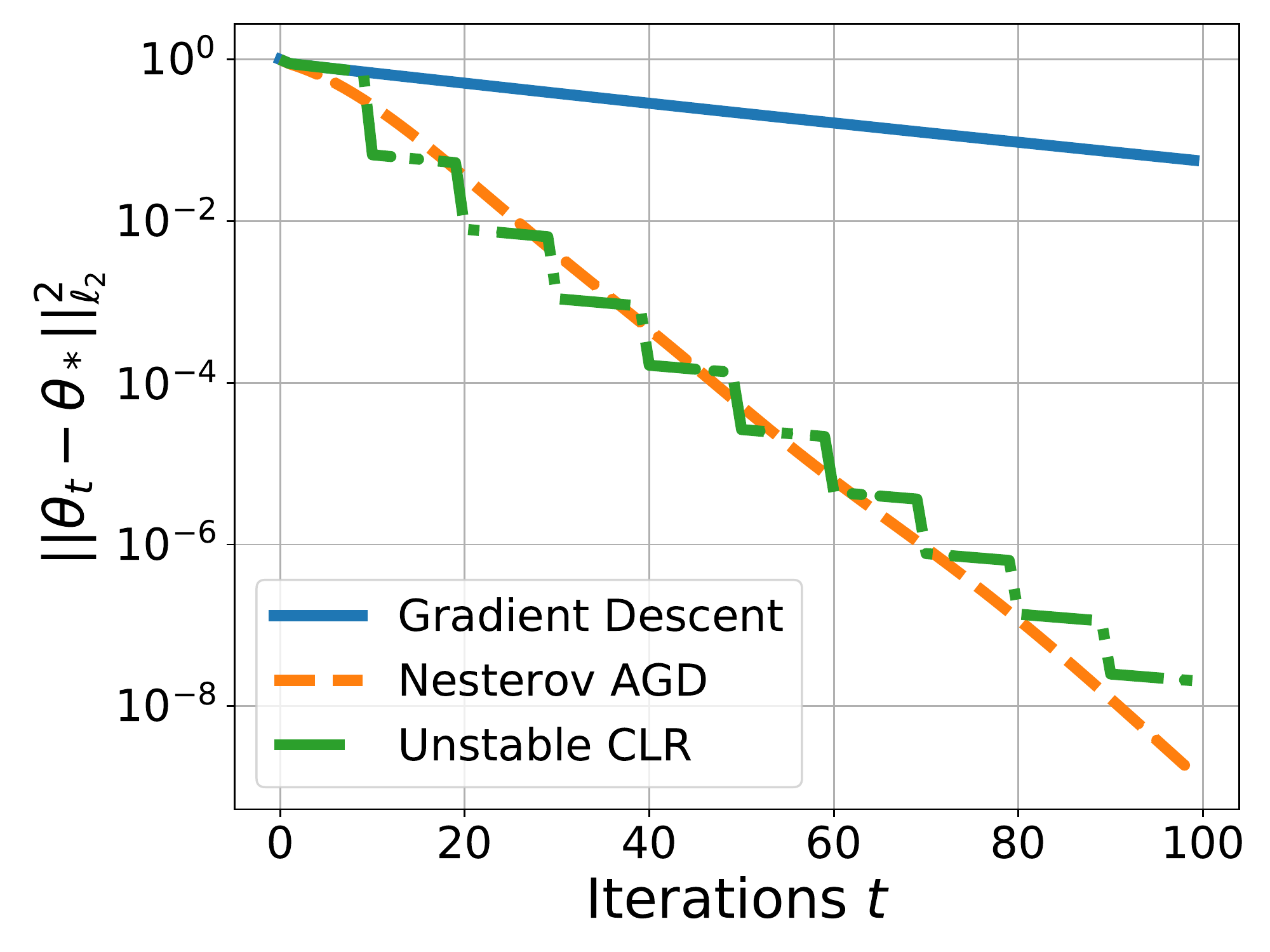}};
\node at (0,-1.75) [scale=0.6] {\Large{(b) $\kappa=100$}};
\end{tikzpicture}\\
\begin{tikzpicture}
\node at (0,0){\includegraphics[width=0.24\textwidth]{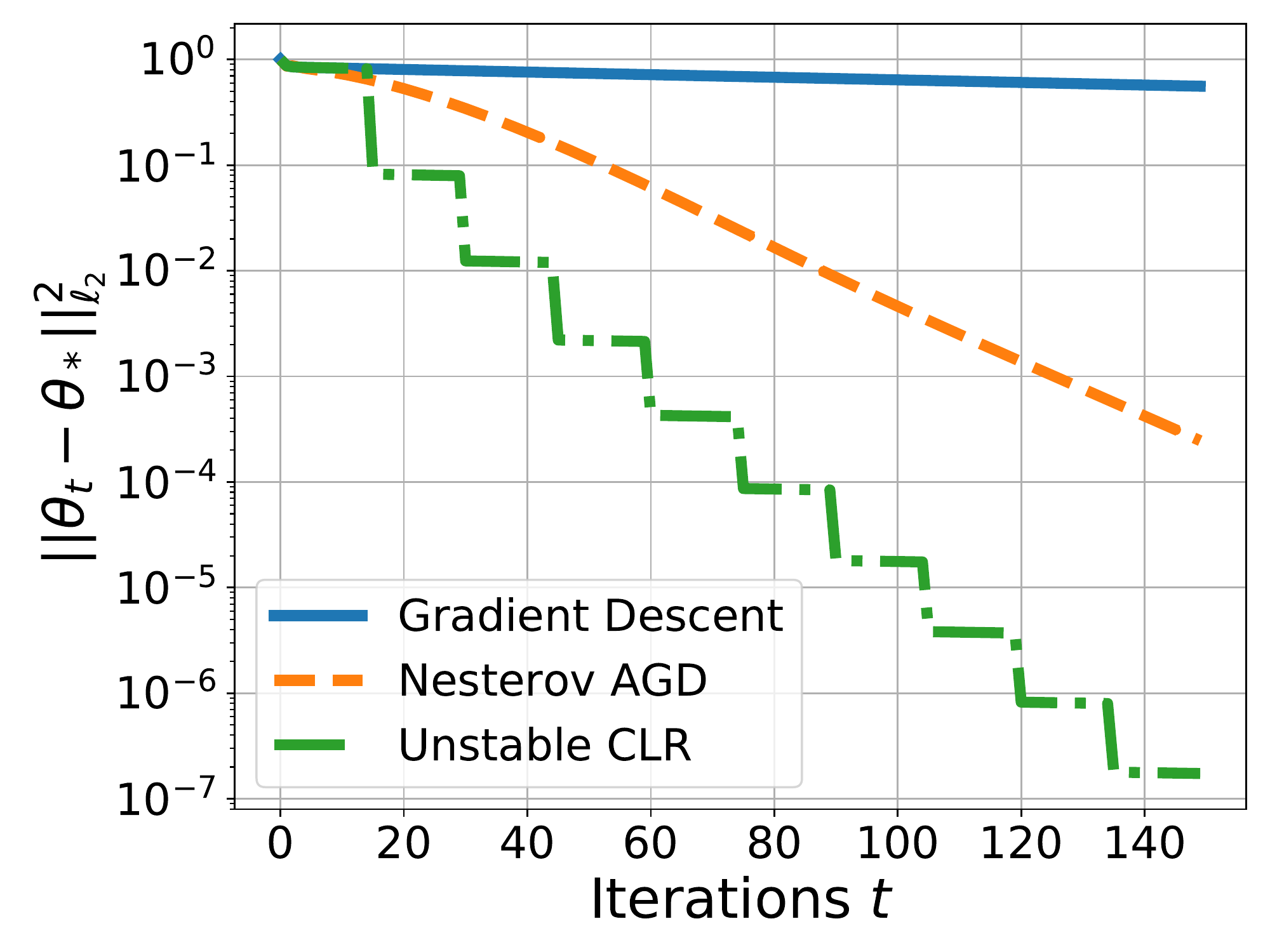}};
\node at (0,-1.75) [scale=0.6] {\Large{(c) $\kappa=1000$}};
\end{tikzpicture}~\hspace{-10pt}
\begin{tikzpicture}
\node at (0,0){\includegraphics[width=0.24\textwidth]{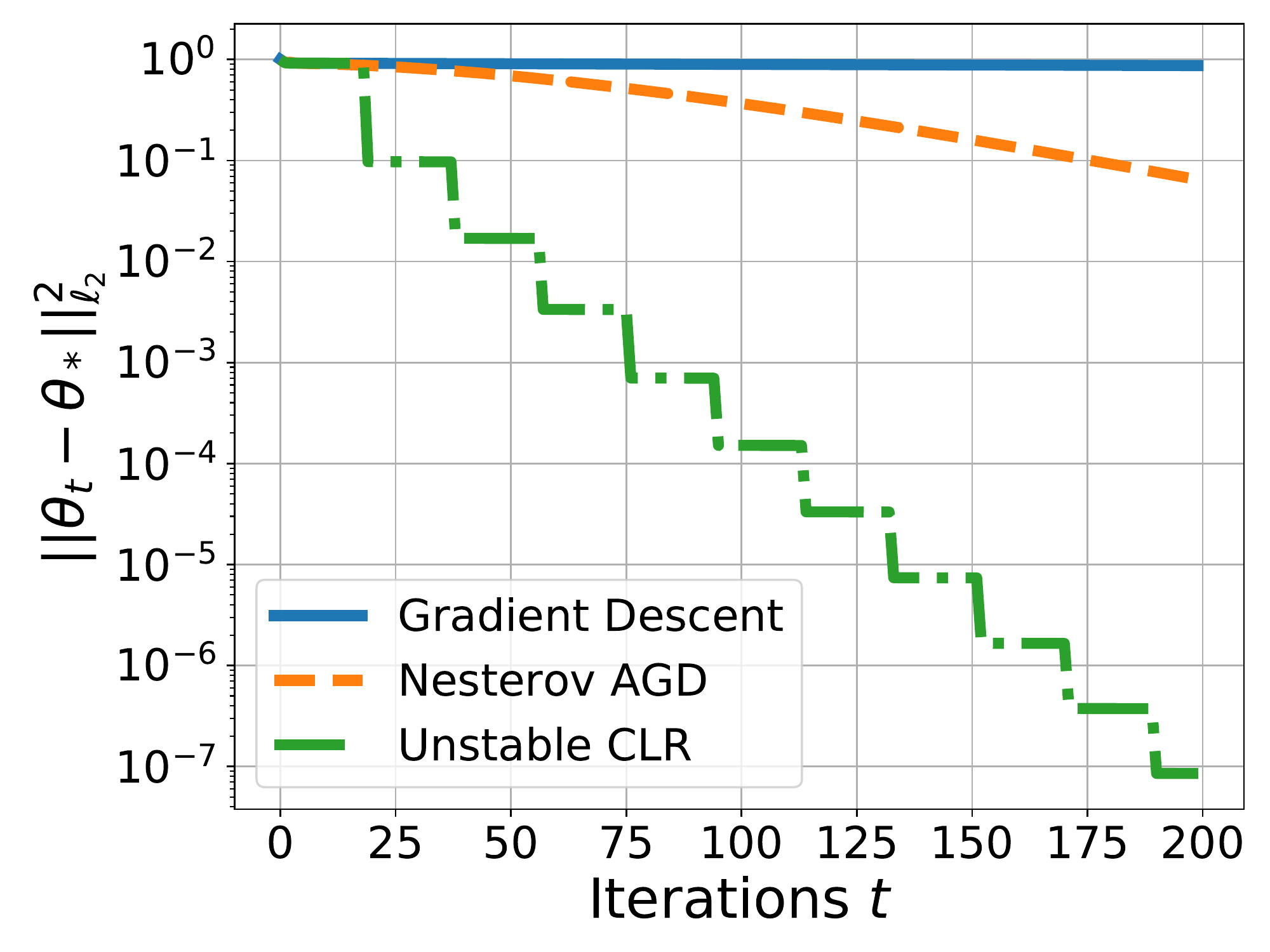}};
\node at (0,-1.75) [scale=0.6] {\Large{(d) $\kappa=10000$}};
\end{tikzpicture}
\caption{Convergence performance of gradient descent with $\eta_t=1/L$, Nesterov's acceleration (AGD) and Unstable Cyclical Learning Rate of Theorem \ref{strcthm} on a linear regression task. In Figures (a) to (d), we vary the condition number $\kappa$ from $10$ to $10^4$. In these experiments, eigenspectrum of Hessian is bimodal where eigenvalues lie over the intervals $[1,2]$ and $[\kappa/2,\kappa]$. This means the local condition numbers are $\kappa_+=\kappa_-=2$. As the global condition number grows, Unstable CLR outperforms \red{standard} gradient descent or Nesterov AGD as it only requires logarithmic iteration in $\kappa$. We note that Unstable CLR can potentially further benefit from acceleration.}\label{fig cond}
\end{figure}

Here, we point out the possibility that, one can use an \red{unstably large} learning rate once in a while to provide \emph{huge improvements over small directions}. The resulting instability can be compensated quickly by following this unstable step with multiple stable steps which keep the larger directions under control. Overall, for problems with bimodal Hessian spectrum, where eigenvalues are clustered in large and small groups, this leads to substantial improvements with logarithmic dependency on the condition number. Figure \ref{fig cond} highlights this phenomena. Our approach can be formalized by using a cyclical (aka periodic) learning rate schedule \cite{smith2017cyclical,loshchilov2016sgdr}. We remark that cyclical learning rate (CLR) \red{is related to SGD with Restarts (SGDR) and Stochastic Weight Averaging} which attracted significant attention due to their fast convergence, generalization benefits and flexibility \cite{smith2017super,fu2019cyclical,izmailov2018averaging}. \cite{li2020exponential} assesses certain theoretical benefits of cosine learning rates which are periodic. \red{\cite{zhang2020global,daneshmand2018escaping} investigate periodic learning rates to facilitate escape from saddle points with stochastic gradients.} Large initial learning rates are also known to play a critical role in generalization \cite{leclerc2020two,krizhevsky2012imagenet,simonyan2014very}. Recent work \cite{edge} provides further empirical evidence that practical learning rates can be at the edge of stability. However to the best of our knowledge, prior works do not consider potential theoretical benefits of \red{unstably large} learning rate choice. The closest work is by Smith and Topin \cite{smith2017super}. Here authors observe that CLR can operate with very large maximum learning rates and converge ``super fast''. We believe this letter provides a rigorous theoretical support for such observations. We show that maximum learning rate can in fact be unstable and it can be much larger than the maximum stable learning rate. We also show that ``super fast'' convergence can be as fast as logarithmic in condition number (which is drastically better than GD or Nesterov AGD) for suitable problems (discussed further below).

Our CLR scheme simply takes two values $\eta_\pm$ as described below.
\begin{definition}\label{def clr} Fix an integer $T>1$ and positive scalars $\eta_+,\eta_-$. Set periodic learning rate $\eta_t$ for $t\geq 0$ as
\[
\eta_t=\begin{cases}\eta_-\quad\text{if}\quad\text{mod}(t,T)=-1\\\eta_+\quad\text{else}\end{cases}.
\]
\end{definition}

We call this schedule \emph{Unstable CLR} if $\eta_->2/L$ where $L$ is the smoothness of the problem. Indeed when this happens, the algorithm is susceptible to blow up in certain eigendirections as the contraction matrix $\Iden-\eta_-\X^\top\X$ has operator norm larger than $1$.

\begin{theorem} [Linear regression] \label{strcthm} Let $\bla\in\R_+^p$ be the decreasingly sorted eigenvalues of $\X^\top\X$\red{~with $L=\la_1$ and $\mu=\la_p$}. Fix some integer $r$ with $p>r\ge1$. Introduce the quantities
\[
\kappa=\frac{L}{\mu},\quad \kp=\frac{\red{L}}{\la_r},\quad \km=\frac{\la_{r+1}}{\red{\mu}}.
\]
Set period $T\geq \kp\log\left(\frac{2\kappa}{2\km-1}\right)+1$ and learning rate according to Def.~\ref{def clr} with $\eta_+=\frac{1}{L}$ and $\eta_-=\frac{1}{\km\mu}$. For all $t$ with $\text{mod}(t,T)=0$, the iterations obey
\begin{align}
\tn{\bt_{t}-\bts}\leq \left(1-\frac{1}{2\km}\right)^{t/T}\tn{\bt_0-\bts}.\label{super fast}
\end{align}
Alternatively, for $t\geq 2T\km\log(\eps^{-1})$, we have $\tn{\bt_{t}-\bts}\leq \eps \tn{\bt_{0}-\bts}$ for $1>\eps>0$.
\end{theorem}\vspace{3pt}
\noindent\textbf{Interpretation.} Observe that in order to achieve $\eps$ accuracy, the number of required iterations grow as 
\begin{align}
\boxed{\kp\km\log\left({\kappa}\right)\log(\eps^{-1}).}\label{req iter}
\end{align}
Here $\kp,\km$ are \red{the} \emph{local condition numbers} for the eigen-spectrum. $\kp$ is the condition number over the subspace $\Sc$ (i.e.~eigens from $\la_1$ to $\la_r$) and $\km$ is the condition number over the orthogonal complement $\Sc^c$. Observe that $\kp\times \km$ is always upper bounded by the overall condition number $\kappa$ i.e.~$\kp\km\leq \kappa$. However if eigenvalues over $\Sc$ and $\Sc^c$ are narrowly clustered, we can have $\kp\km\ll \kappa$ leading to a much faster convergence. Finally observe that the dependence on the (global) condition number $\kappa$ \red{is} only logarithmic. \red{Thus, in the worst case scenario of $\kappa_+\kappa_-=\kappa$, the iteration complexity \eqref{req iter} is sub-optimal by a $\log(\kappa)$-factor compared to the standard gradient descent which requires $\kappa\log(\eps^{-1})$ iterations}. \red{However there is a factor of $\kappa/\log(\kappa)$ improvement when the eigenvalues are clustered and $\kappa_+,\kappa_-$ are small.} Figure \ref{fig cond} demonstrates the comparisons to gradient descent (with $\eta=1/L$) and Nesterov's Accelerated GD. In these examples, we set local condition numbers to $\km=\kp=2$ whereas $\kappa$ varies from $10$ to $10000$. As $\kappa$ grows larger, Unstable CLR shines over the alternatives as the iteration number only grows as $\log(\kappa)$. \red{Finally, note that inverse learning rate $\eta_+^{-1}=L$ is equal to the smoothness (top Hessian eigenvalue) of the whole problem whereas $\eta_-^{-1}=\kappa_-\mu$ is chosen based on the smoothness over the lower spectrum.}

\vspace{3pt}
\noindent\textbf{Bimodal Hessian.} The bimodal Hessian spectrum is crucial for enabling super fast convergence. In essence, with bimodal structure, Unstable CLR acts as a \emph{preconditioner} for the problem and helps not only large eigen-value/directions but also small ones. It is possible that other cyclical schemes will accelerate a broader class of Hessian spectrums. That said, we briefly mention that bimodal spectrum has empirical support in the deep learning literature. Several works studied the empirical Hessians (and Jacobians) of deep neural networks \cite{sagun2017empirical,oymak2019generalization,li2020hessian}. A common observation is that the Hessian spectrum has relatively few large eigenvalues and many more smaller eigenvalues \cite{sagun2017empirical,li2020hessian}. \red{These observations are closely related to the seminal \emph{spiked covariance model} \cite{paul2007asymptotics} where the covariance spectrum has a few large eigenvalues and many more smaller eigenvalues.} In connection, \cite{oymak2019generalization,li2020gradient,kopitkov2020neural} studied the Jacobian spectrum. Note that, in the linear regression setting of Theorem \ref{strcthm}, Jacobian is simply $\X$. They similarly found that practical deep nets have a bimodal spectrum and the Jacobian is approximately low-rank. It is also known that behavior of the wide deep nets (i.e.,~many neurons per layer) can be approximated by their Jacobian-based linearization at initialization via neural tangent kernel \cite{jacot2018neural,lee2019wide}. Thus the spectrum indeed closely governs the optimization dynamics \cite{gur2018gradient} similar to our simple linear regression setup. While the larger eigendirections of a deep net can be optimized quickly \cite{oymak2019generalization,kopitkov2020neural}, intuitively their existence will slow down the small eigendirections. However it is also observed that learning such small eigendirections is critical for success of deep learning since typically achieving zero-training loss (aka interpolation) leads to the best generalization performance \cite{belkin2019reconciling,belkin2018overfitting,poggio2017theory}. In light of this discussion, related empirical observations of \cite{smith2017super} and our theory, it is indeed plausible that Unstable CLR does a stellar job at learning these small eigen-directions leading to much faster interpolation and improved generalization.



\section{Extension to the Nonlinear Problems}
To conclude with our discussion, we next provide a more general result that apply to strongly-convex functions. Recall $\Sc^c$ as the complement of a subspace $\Sc$. Let $\Pis$ denote the matrix that projects onto $\Sc$. Our goal is solving $\bts=\arg\min_{\bt\in\R^p}f(\bt)$ via gradient iterations $\bt_{t+1}=\bt_t-\eta_t\gradf{\bt_t}$.
\begin{definition}[Smoothness and strong-convexity] \label{ssc}Let $f:\R^p\rightarrow\R$ be a smooth convex function and fix $L>\mu>0$. $f:\R^p\rightarrow\R$ is $L$-smooth and $\mu$ strongly-convex if its Hessian satisfies the following for all $\bt\in\R^p$
\[
L\Iden_p\succeq \hessf{\bt}\succeq \mu\Iden_p.
\]
\end{definition}
First, we recall a classical convergence result for strongly convex functions for the sake of completeness.
\begin{proposition} \label{strcthm}Let $f$ obey Definition \ref{ssc} and suppose $\bts$ is its minimizer. Pick a learning rate $\eta\leq 1/L$, and use the iterations $\bt_{t+1}=\bt_t-\eta\gradf{\bt_t}$. The iterates obey $\tn{\bt_{\tau}-\bts}^2\leq (1-\eta\mu)^{\tau}\tn{\bt_{0}-\bts}^2$.
\end{proposition}
This is the usual setup for gradient descent analysis. Instead, we will employ the bimodal Hessian definition.
\begin{definition}[Bimodal Hessian] \label{dssc}Let $f:\R^p\rightarrow\R$ be an $L$ smooth $\mu$ strongly-convex function. Additionally, there exists a subspace $\Sc\in\R^p$ and local condition numbers $\kp,\km\geq 1$ and cross-smoothness $\eps\geq 0$ such that, the Hessian of $f$ is satisfies\vspace{-2pt}
\begin{align}\label{hesseq}
&\text{Upper spectrum:}~~~L\Iden_p\succeq \Pis\hessf{\bt}\Pis\succeq (L/\kp)\Iden_p, \\
&\text{Lower spectrum:}~~~\km\mu\Iden_p\succeq \Pisc\hessf{\bt}\Pisc\succeq \mu\Iden_p, \nn\\
&\text{Cross spectrum:}~~~\|\Pis\hessf{\bt}\Pisc\|\leq \eps.\nn
\end{align}
\end{definition}
Here $\kp,\km$ are the local condition numbers as previously. Observe that both of these are upper bounded by the global condition number $\kappa=L/\mu$ as $f$ obeys Def.~\ref{ssc} as well. The cross-smoothness controls the interaction between two subspaces. For linear regression $\eps=0$ by picking $\Sc$ to be eigenspace. For general nonlinear models, as long as problem can be approximated linearly (e.g.~wide deep nets can be approximated by their linearization \cite{jacot2018neural,lee2019wide}), it is plausible that cross-smoothness is small for a suitable choice of $\Sc$. We have the following analogue of Theorem \ref{strcthm}. 
\begin{theorem} [Nonlinear problems] \label{thm hessian}Let $f$ obey Definition \ref{dssc} with non-negative scalars $L,\mu,\kp,\km,\eps$. Consider the learning rate schedule of Definition~\ref{def clr} and set
\[
T\geq2\kp\log\left(\frac{2L}{\km \mu}\right)+1\quad,\quad\eta_-=\frac{1}{\km \mu}\quad,\quad\eta_+=\frac{1}{L}.
\]
Additionally, suppose the cross-smoothness $\eps$ satisfies the following upper bound
\[
4\eps\leq \min(1,{\km}/{T})~\mu.
\]
Let $\bts$ be the minimizer of $f(\bt)$. Starting from $\bt_0$, apply the gradient iterations $\bt_{t+1}=\bt_t-\eta_t\gradf{\bt_t}$. For all iterations $t\geq 0$ with $\text{mod}(t,T)=0$, we have
\[
\tn{\bt_{t}-\bts}\leq \sqrt{2}(1-\frac{1}{4\km})^{t/T}\tn{\bt_0-\bts}.
\]
\end{theorem}
\noindent\textbf{Interpretation.} This result is in similar spirit to the linear regression setup of Theorem \ref{strcthm}. The required number of iterations is still governed by the quantity \eqref{req iter}. A key difference is the cross-smoothness $\eps$ which controls the cross-Hessian matrix $\Pis\hessf{\bt}\Pisc$. This term was simply equal to $0$ for the linear problem. In essence, our condition for nonlinear problems essentially requires cross-Hessian to be dominated by the Hessian over the lower spectrum i.e.~$\Pisc\hessf{\bt}\Pisc$. In particular, $\eps$ should be upper bounded by the strong convexity parameter $\mu$ as well as ${\km\mu}/{T}$ where $\km\mu$ is the smoothness over the lower spectrum. It would be interesting to explore to what extent the conditions on the cross-Hessian can be relaxed. However, as mentioned earlier, the fact that wide artificial neural networks behave close to linear models \cite{jacot2018neural} provides a decent justification for small cross-Hessian. 

\section{Proofs}
\subsection{Proof of Theorem \ref{strcthm}}
\begin{proof} Following Theorem \ref{strcthm}'s statement introduce $L=\la_1,~\mu=\la_p$. Each gradient iteration can be written in terms of the residual $\w_t=\bt_t-\bts$ and has the following form
\begin{align}
&\bt_{t+1}=\bt_t-\eta_t\X^T(\y-\X\bt_t)\nn\\
&\implies \w_{t+1}:=\bt_{t+1}-\bts=(\Iden-\X^T\X)\w_t.\nn
\end{align}
Let $\Sc$ be the principal eigenspace induced by the first $r$ eigenvectors and $\Sc^c$ be its complement. Let $\Pis$ be the projection operator on $\Sc$. Set $\ab_t=\Pis(\w_t)$, $\bb_t=\Pisc(\w_t)$. Also set $a_t=\tn{\ab_t}$, $b_t=\tn{\bb_t}$. Since the learning rate is periodic, we analyze a single period starting from $\ab_0,\bb_0$. During the first $T-1$ iterations, we have that
\begin{align}
a_t\leq (1-\frac{1}{\kp})^t a_0\quad\text{and}\quad b_t\leq (1-\frac{\mu}{L})^t b_0=(1-\frac{1}{\kappa})^t b_0.\nn
\end{align}
At the final (\red{unstably large}) iteration $t=T-1$, we have
\begin{align}
a_T\leq \frac{L}{\km\mu} a_{T-1}= \frac{\kappa}{\km} a_{T-1} \quad\text{and}\quad b_t\leq (1-\frac{1}{\km}) b_0.\nn
\end{align}
Now $b_t$ term is clearly non-increasing and obeys
$
b_T\leq (1-\frac{1}{\km})(1-\frac{1}{\kappa})^{T-1}b_0\leq (1-\frac{1}{\km})b_0.
$
Note that we forego the $(1-\frac{1}{\kappa})^{T-1}$ for the sake of simplicity. We wish to make the growth of $a_T$ similarly small by enforcing
\begin{align}
&\frac{\kappa}{\km}(1-\frac{1}{\kp})^{T-1}\leq 1-\frac{1}{2\km}\iff\frac{2\kappa}{2\km-1}\leq ((1-\frac{1}{\kp})^{-1})^{T-1}.\nn
\end{align}
Observe that $(1-\frac{1}{\kp})^{-1}\geq \e^{1/\kp}$. Thus, we simply need
\[
\log(\frac{2\kappa}{2\km-1})\leq \frac{T-1}{\kp}\iff T\geq \kp{\log(\frac{2\kappa}{2\km-1})}+1.
\]
In short, at the end of a single period we are guaranteed to have \eqref{super fast} after observing $\tn{\w_t}^2=a_t^2+b_t^2$.
\end{proof}
\subsection{Proof of Theorem \ref{thm hessian}}
\begin{proof} The proof idea is same as Theorem \ref{strcthm} but we additionally control the cross-smoothness terms. Denote the residual by $\w_t=\bt_t-\bts$. We will work with the projections of the residual on the subspaces $\Sc,\Sc^c$ denoted by $\ab_t,\bb_t$ respectively. Additionally, set $B=\max(\tn{\ab_0},\tn{\bb_0})$ and set $a_t=\tn{\ab_t}/B,b_t=\tn{\bb_t}/B$. Observe that by this definition $B\leq \tn{\bts-\bt_0}$ and
\[
\max(a_0,b_0)=1.
\]
We will prove that at the end of a single period, $(a_T,b_T)$ pair obeys
\begin{align}
\max(a_T,b_T)\leq 1-\frac{1}{4\km}.\label{key claim}
\end{align}
The overall result follows inductively from this as follows. First, inductively we would achieve $\max(a_t,b_t)\leq (1-\frac{1}{4\km})^{t/T}$ for $\text{mod}(t,T)=0$. That would in turn yield
\begin{align}
\tn{\bt_{t}-\bts}&\leq \sqrt{2}\max(\tn{\ab_t},\tn{\bb_t})\leq \sqrt{2}B\max(a_t,b_t)\nn\\
&\leq \sqrt{2}(1-\frac{1}{4\km})^{t/T}\tn{\bts-\bt_0}.\nn
\end{align}
\noindent\textbf{Establishing \eqref{key claim}:} Thus, let us show \eqref{key claim} for a single period of learning rate i.e.~$0\leq t\leq T$.
The gradient is given by
\[
\gradf{\bt_t}=\Hb\w_t,
\]
where $\Hb$ is obtained by integrating the Hessian's along the path from $\bts$ to $\bt_t$. Write the next iterate as
\[
\tn{\w_{t+1}}^2=\tn{\w_t-\eta_t\gradf{\bt_t}}^2=\tn{\w_t-\eta_t\Hb\w_t}^2.
\]
By Definition \ref{dssc} and triangle inequality, $\Hb$ satisfies the bimodal Hessian inequalities described in \eqref{hesseq}. 
To analyze a single period, we first focus on the lower learning rate $\eta_+$ which spans the initial $T-1$ iterations. Hence, suppose $\eta_t=\eta_+= 1/L$. Note that from strong convexity of $\Hb$ over $\Sc,\Sc^c$, we have that
\begin{align*}
&\tn{\ab_t-\eta_t\Pis\Hb\ab_t}^2\leq (1-\eta_tL/\kp)\tn{\ab_t}^2\quad\text{for}\quad \eta_t\leq 1/L\\
&\tn{\bb_t-\eta_t\Pisc\Hb\bb_t}^2\leq (1-\eta_t\mu)\tn{\bb_t}^2\quad\quad\text{for}\quad \eta_t\leq 1/(\km \mu).
\end{align*}
Following this with $\eta_t=\eta_+=1/L$, for $\ab_{t}$, we find the recursions 
\begin{align}
\tn{\ab_{t+1}}&=\tn{\ab_t-\eta_t\Pis\Hb\w_t}\nn\\
&\leq \tn{\ab_t-\eta_t\Pis\Hb\ab_t}+\eta_t\tn{\Pis\Hb\bb_t}\nn\\
&\leq (1-\frac{\eta_tL}{2\kp})\tn{\ab_t}+\eta_t\eps\tn{\bb_t}\nn\\
&=(1-\frac{1}{2\kp})\tn{\ab_t}+\frac{\eps}{L}\tn{\bb_t}.\label{core arg}
\end{align}
Using the identical argument for $\bb_t$ yields
\begin{align}
\tn{\bb_{t+1}}&= \tn{\bb_t-\eta_t\Pisc\Hb\w_t}\nn\\
&\leq \tn{\bb_t-\eta_t\Pisc\Hb\bb_t}+\eta_t\tn{\Pisc\Hb\ab_t}\nn\\
&\leq (1-\frac{\mu}{2L})\tn{\bb_t}+\frac{\eps}{ L}\tn{\ab_t}.\nn
\end{align}
Setting $\bar{\eps}=\frac{\eps}{ L}$, we obtain\vspace{-10pt}
\begin{align}
&a_{t+1}\leq (1-\frac{1}{2\kp})a_t+\bar{\eps}b_t\label{a bound}.
\end{align}
Additionally, recall that, since $f$ is $\mu$ strongly-convex we have $L/\kp\geq \mu$. Thus using $L/\kp\geq\mu\geq 2\eps$ we have
\begin{align*}
&\tn{\bb_{t+1}}\leq (1-\frac{\mu}{2L})\tn{\bb_t}+\bar{\eps}\tn{\ab_t}\leq \max(\tn{\ab_t},\tn{\bb_t})\\
&\tn{\ab_{t+1}}\leq (1-\frac{1}{2\kp})\tn{\ab_t}+\bar{\eps}\tn{\bb_t}\leq \max(\tn{\ab_t},\tn{\bb_t}).
\end{align*}
That is, we are guaranteed to have $1\geq a_t,b_t\geq 0$. Thus, using \eqref{a bound}, recursively, for all $0\leq t\leq T-1$, $a_t$ satisfies 
\begin{align}
a_t& \leq (1-\frac{1}{2\kp})^t +\bar{\eps}\sum_{\tau=0}^{t-1}(1-\frac{1}{2\kp})^{t-\tau-1} b_t\leq (1-\frac{1}{2\kp})^t+\bar{\eps}\sum_{\tau=0}^{t-1} b_\tau\nn\\
&\leq (1-\frac{1}{2\kp})^t+t\bar{\eps}\label{arec}.
\end{align}
At time $t=T-1$, we use the larger learning rate $\eta_-= \frac{1}{\km\mu}$. The following holds via identical argument as \eqref{core arg}
\begin{align}
&a_{t+1}\leq \frac{L}{\km\mu}a_t+\frac{\eps}{\km\mu}b_t\label{a cond large}\\
&b_{t+1}\leq (1-\frac{1}{2\km})b_t+\frac{\eps}{\km\mu}a_t.\nn
\end{align}
To bound $b_T$ at time $t=T-1$, we recall $b_{T-1},a_{T-1}\leq 1$ and the bound $\eps\leq \mu/4$, to obtain
\[
b_T\leq 1-\frac{1}{2\km}+\frac{\eps}{\km\mu}\leq 1-\frac{1}{4\km}.
\]
Bounding $a_T$ is what remains. Noticing $\log(1-x)\leq {-x}$, our period choice $T$ obeys
\[
T\geq 2\kp{\log(2L/(\km\mu))}+1\geq  1-\frac{\log(2L/(\km\mu))}{\log({1-\frac{1}{2\kp}})}
\]
for $\kp\geq 1$. Thus, using the assumption $\eps\leq \frac{\km\mu}{4T}$, and the bounds \eqref{arec} and \eqref{a cond large}, the bound on $a_T$ is obtained via
\begin{align}
a_T&\leq \frac{L}{\km\mu}((1-\frac{1}{2\kp})^{T-1}+(T-1)\frac{\eps}{L})+\frac{\eps}{\km\mu}\nn\\
&\leq \frac{1}{2}+T\frac{\eps}{\km\mu}\leq \frac{3}{4}\leq1-\frac{1}{4\km},\nn
\end{align}
where we noted $\km\geq1$. The above bounds on $a_T,b_T$ establishes \eqref{key claim} and concludes the proof.
\end{proof}

\section{Conclusions}

This letter introduced a setting where remarkably fast convergence can be attained by using cyclical learning rate schedule that carefully targets the bimodal structure of the Hessian of the problem. This is accomplished by replacing global condition number with local counterparts. While bimodal Hessian seems to be a strong assumption, recent literature provides rich empirical justification for our theory. Besides relaxing the assumptions in Theorem \ref{thm hessian}, there are \red{multiple interesting directions for Unstable CLR. The results can likely be extended to minibatch SGD, overparameterized settings}, or to non-convex problems (e.g.~via Polyak-Lojasiewicz condition \cite{karimi2016linear}). \red{While our attention was restricted to bimodal Hessian and a CLR scheme with two values ($\eta_+,\eta_-$ as in Def.~\ref{def clr}), it would be exciting to explore whether more sophisticated CLR schemes can provably accelerate optimization for a richer class of Hessian structures.} On the practical side, further empirical investigation (beyond \cite{smith2017super,edge}) can help verify and explore the potential benefits of \red{large cyclical} learning rates.

\small{
\bibliographystyle{IEEEtran}
\bibliography{Bibfiles}
}

\end{document}

%% file: notation.tex
\usepackage{hyperref,graphicx,amsmath,amsfonts,amssymb,bm,url,breakurl,epsfig,epsf,color,MnSymbol,mathbbol,cite,caption,subcaption,multirow,comment}
\usepackage[noend]{algpseudocode}

\usepackage{amssymb}

\usepackage[utf8]{inputenc} 
\usepackage[T1]{fontenc}    
\usepackage{url}            
\usepackage{booktabs}       
\usepackage{amsfonts}       
\usepackage{nicefrac}       
\usepackage{microtype}      

\makeatletter
\providecommand*{\boxast}{%
  \mathbin{
    \mathpalette\@boxit{*}%
  }%
}
\makeatother

  \makeatletter
\def\BState{\State\hskip-\ALG@thistlm}
\makeatother

  \usepackage{mathtools}

\usepackage{tikz}
\usepackage{pgfplots}
\usetikzlibrary{pgfplots.groupplots}

\usepackage{movie15}

\usepackage{cite}
\usepackage{caption}

\newcommand{\tsn}[1]{{\left\vert\kern-0.25ex\left\vert\kern-0.25ex\left\vert #1 
    \right\vert\kern-0.25ex\right\vert\kern-0.25ex\right\vert}}

\definecolor{darkred}{RGB}{150,0,0}
\definecolor{darkgreen}{RGB}{0,150,0}
\definecolor{darkblue}{RGB}{0,0,200}
\hypersetup{colorlinks=true, linkcolor=darkred, citecolor=darkgreen, urlcolor=darkblue}

\newtheorem{theorem}{Theorem}

\newtheorem{proposition}{Proposition}
\newtheorem{definition}{Definition}

\newcommand{\eps}{\varepsilon}

\newcommand{\Pis}{\boldsymbol{\Pi}_\Sc}
\newcommand{\Pisc}{\boldsymbol{\Pi}_{\Sc^c}}

\newcommand{\beq}{\begin{equation}}

\newcommand{\eeq}{\end{equation}}

\newcommand{\nn}{\nonumber}
\newcommand{\la}{\lambda}
\newcommand{\km}{\kappa_-}

\newcommand{\kp}{\kappa_+}
\newcommand{\bla}{\boldsymbol{\lambda}}

\newcommand{\bt}{\bteta}
\newcommand{\bts}{\bteta_\star}

\newcommand{\Cb}{{\mtx{C}}}

\newcommand{\Hb}{{\mtx{H}}}

\newcommand{\red}[1]{{\textcolor{black}{#1}}}

\newcommand{\Iden}{{\mtx{I}}}

\newcommand{\smn}[1]{{\sigma_{\min}(#1)}}

\newcommand{\tn}[1]{\|{#1}\|_{\ell_2}}

\newcommand{\bteta}{\boldsymbol{\theta}}

\newcommand{\Sc}{\mathcal{S}}

\newcommand{\w}{\vct{w}}

\newcommand{\ab}{\vct{a}}
\newcommand{\bb}{\vct{b}}

\newcommand{\y}{\vct{y}}

\definecolor{emmanuel}{RGB}{255,127,0}

\newcommand{\R}{\mathbb{R}}

\newcommand{\gradf}[1]{{\boldsymbol{\nabla} f(#1)}}
\newcommand{\hessf}[1]{{\boldsymbol{\nabla}^2f(#1)}}

\newcommand{\e}{\mathrm{e}}

\newcommand{\vct}[1]{\bm{#1}}
\newcommand{\mtx}[1]{\bm{#1}}

\newcommand{\X}{{\mtx{X}}}

\def \endprf{\hfill {\vrule height6pt width6pt depth0pt}\medskip}
\newenvironment{proof}{\noindent {\bf Proof} }{\endprf\par}

%% file: OymakSPL.bbl
\begin{thebibliography}{10}
\providecommand{\url}[1]{#1}
\csname url@samestyle\endcsname
\providecommand{\newblock}{\relax}
\providecommand{\bibinfo}[2]{#2}
\providecommand{\BIBentrySTDinterwordspacing}{\spaceskip=0pt\relax}
\providecommand{\BIBentryALTinterwordstretchfactor}{4}
\providecommand{\BIBentryALTinterwordspacing}{\spaceskip=\fontdimen2\font plus
\BIBentryALTinterwordstretchfactor\fontdimen3\font minus
  \fontdimen4\font\relax}
\providecommand{\BIBforeignlanguage}[2]{{%
\expandafter\ifx\csname l@#1\endcsname\relax
\typeout{** WARNING: IEEEtran.bst: No hyphenation pattern has been}%
\typeout{** loaded for the language `#1'. Using the pattern for}%
\typeout{** the default language instead.}%
\else
\language=\csname l@#1\endcsname
\fi
#2}}
\providecommand{\BIBdecl}{\relax}
\BIBdecl

\bibitem{smith2017cyclical}
L.~N. Smith, ``Cyclical learning rates for training neural networks,'' in
  \emph{Applications of Computer Vision (WACV), 2017 IEEE Winter Conference
  on}.\hskip 1em plus 0.5em minus 0.4em\relax IEEE, 2017, pp. 464--472.

\bibitem{loshchilov2016sgdr}
I.~Loshchilov and F.~Hutter, ``Sgdr: Stochastic gradient descent with warm
  restarts,'' \emph{arXiv preprint arXiv:1608.03983}, 2016.

\bibitem{smith2017super}
L.~N. Smith and N.~Topin, ``Super-convergence: Very fast training of residual
  networks using large learning rates,'' \emph{arXiv preprint
  arXiv:1708.07120}, 2017.

\bibitem{fu2019cyclical}
H.~Fu, C.~Li, X.~Liu, J.~Gao, A.~Celikyilmaz, and L.~Carin, ``Cyclical
  annealing schedule: A simple approach to mitigating kl vanishing,''
  \emph{arXiv preprint arXiv:1903.10145}, 2019.

\bibitem{izmailov2018averaging}
P.~Izmailov, D.~Podoprikhin, T.~Garipov, D.~Vetrov, and A.~G. Wilson,
  ``Averaging weights leads to wider optima and better generalization,'' in
  \emph{34th Conference on Uncertainty in Artificial Intelligence 2018, UAI
  2018}.\hskip 1em plus 0.5em minus 0.4em\relax Association For Uncertainty in
  Artificial Intelligence (AUAI), 2018, pp. 876--885.

\bibitem{li2020exponential}
X.~Li, Z.~Zhuang, and F.~Orabona, ``Exponential step sizes for non-convex
  optimization,'' \emph{arXiv preprint arXiv:2002.05273}, 2020.

\bibitem{zhang2020global}
K.~Zhang, A.~Koppel, H.~Zhu, and T.~Basar, ``Global convergence of policy
  gradient methods to (almost) locally optimal policies,'' \emph{SIAM Journal
  on Control and Optimization}, vol.~58, no.~6, pp. 3586--3612, 2020.

\bibitem{daneshmand2018escaping}
H.~Daneshmand, J.~Kohler, A.~Lucchi, and T.~Hofmann, ``Escaping saddles with
  stochastic gradients,'' in \emph{International Conference on Machine
  Learning}.\hskip 1em plus 0.5em minus 0.4em\relax PMLR, 2018, pp. 1155--1164.

\bibitem{leclerc2020two}
G.~Leclerc and A.~Madry, ``The two regimes of deep network training,''
  \emph{arXiv preprint arXiv:2002.10376}, 2020.

\bibitem{krizhevsky2012imagenet}
A.~Krizhevsky, I.~Sutskever, and G.~E. Hinton, ``Imagenet classification with
  deep convolutional neural networks,'' \emph{Advances in neural information
  processing systems}, vol.~25, pp. 1097--1105, 2012.

\bibitem{simonyan2014very}
K.~Simonyan and A.~Zisserman, ``Very deep convolutional networks for
  large-scale image recognition,'' \emph{arXiv preprint arXiv:1409.1556}, 2014.

\bibitem{edge}
J.~M. Cohen, S.~Kaur, Y.~Li, Z.~Kolter, and A.~Talwalkar, ``Gradient descent on
  neural networks typically occurs at the edge of stability,'' \emph{to appear
  at ICLR}, 2021.

\bibitem{sagun2017empirical}
L.~Sagun, U.~Evci, V.~U. Guney, Y.~Dauphin, and L.~Bottou, ``Empirical analysis
  of the hessian of over-parametrized neural networks,'' \emph{arXiv preprint
  arXiv:1706.04454}, 2017.

\bibitem{oymak2019generalization}
S.~Oymak, Z.~Fabian, M.~Li, and M.~Soltanolkotabi, ``Generalization guarantees
  for neural networks via harnessing the low-rank structure of the jacobian,''
  \emph{arXiv preprint arXiv:1906.05392}, 2019.

\bibitem{li2020hessian}
X.~Li, Q.~Gu, Y.~Zhou, T.~Chen, and A.~Banerjee, ``Hessian based analysis of
  sgd for deep nets: Dynamics and generalization,'' in \emph{Proceedings of the
  2020 SIAM International Conference on Data Mining}.\hskip 1em plus 0.5em
  minus 0.4em\relax SIAM, 2020, pp. 190--198.

\bibitem{paul2007asymptotics}
D.~Paul, ``Asymptotics of sample eigenstructure for a large dimensional spiked
  covariance model,'' \emph{Statistica Sinica}, pp. 1617--1642, 2007.

\bibitem{li2020gradient}
M.~Li, M.~Soltanolkotabi, and S.~Oymak, ``Gradient descent with early stopping
  is provably robust to label noise for overparameterized neural networks,'' in
  \emph{International Conference on Artificial Intelligence and
  Statistics}.\hskip 1em plus 0.5em minus 0.4em\relax PMLR, 2020, pp.
  4313--4324.

\bibitem{kopitkov2020neural}
D.~Kopitkov and V.~Indelman, ``Neural spectrum alignment: Empirical study,'' in
  \emph{International Conference on Artificial Neural Networks}.\hskip 1em plus
  0.5em minus 0.4em\relax Springer, 2020, pp. 168--179.

\bibitem{jacot2018neural}
A.~Jacot, F.~Gabriel, and C.~Hongler, ``Neural tangent kernel: Convergence and
  generalization in neural networks,'' \emph{Conference on Neural Information
  Processing Systems}, 2018.

\bibitem{lee2019wide}
J.~Lee, L.~Xiao, S.~S. Schoenholz, Y.~Bahri, R.~Novak, J.~Sohl-Dickstein, and
  J.~Pennington, ``Wide neural networks of any depth evolve as linear models
  under gradient descent,'' \emph{arXiv preprint arXiv:1902.06720}, 2019.

\bibitem{gur2018gradient}
G.~Gur-Ari, D.~A. Roberts, and E.~Dyer, ``Gradient descent happens in a tiny
  subspace,'' \emph{arXiv preprint arXiv:1812.04754}, 2018.

\bibitem{belkin2019reconciling}
M.~Belkin, D.~Hsu, S.~Ma, and S.~Mandal, ``Reconciling modern machine-learning
  practice and the classical bias--variance trade-off,'' \emph{Proceedings of
  the National Academy of Sciences}, vol. 116, no.~32, pp. 15\,849--15\,854,
  2019.

\bibitem{belkin2018overfitting}
M.~Belkin, D.~Hsu, and P.~Mitra, ``Overfitting or perfect fitting? risk bounds
  for classification and regression rules that interpolate,'' \emph{arXiv
  preprint arXiv:1806.05161}, 2018.

\bibitem{poggio2017theory}
T.~Poggio, K.~Kawaguchi, Q.~Liao, B.~Miranda, L.~Rosasco, X.~Boix, J.~Hidary,
  and H.~Mhaskar, ``Theory of deep learning iii: explaining the non-overfitting
  puzzle,'' \emph{arXiv preprint arXiv:1801.00173}, 2017.

\bibitem{karimi2016linear}
H.~Karimi, J.~Nutini, and M.~Schmidt, ``Linear convergence of gradient and
  proximal-gradient methods under the polyak-{\l}ojasiewicz condition,'' in
  \emph{Joint European Conference on Machine Learning and Knowledge Discovery
  in Databases}.\hskip 1em plus 0.5em minus 0.4em\relax Springer, 2016, pp.
  795--811.

\end{thebibliography}
